# A Bayesian hybrid method for context-sensitive spelling correction

**Andrew R. Golding**

Mitsubishi Electric Research Labs

201 Broadway, 8th Floor

Cambridge, MA 02139

golding@merl.com

## Abstract

Two classes of methods have been shown to be useful for resolving lexical ambiguity. The first relies on the presence of particular words within some distance of the ambiguous target word; the second uses the pattern of words and part-of-speech tags around the target word. These methods have complementary coverage: the former captures the lexical "atmosphere" (discourse topic, tense, etc.), while the latter captures local syntax. Yarowsky has exploited this complementarity by combining the two methods using decision lists. The idea is to pool the evidence provided by the component methods, and to then solve a target problem by applying the single strongest piece of evidence, whatever type it happens to be. This paper takes Yarowsky's work as a starting point, applying decision lists to the problem of context-sensitive spelling correction. Decision lists are found, by and large, to outperform either component method. However, it is found that further improvements can be obtained by taking into account not just the single strongest piece of evidence, but *all* the available evidence. A new hybrid method, based on Bayesian classifiers, is presented for doing this, and its performance improvements are demonstrated.

## 1 Introduction

Two classes of methods have been shown useful for resolving lexical ambiguity. The first tests for the presence of particular *context words* within a certain distance of the ambiguous target word. The second tests for *collocations* — patterns of words and part-of-speech tags around the target word. The context-word and collocation methods have complementary coverage: the former captures the lexical "atmosphere" (discourse topic, tense, etc.), while the latter captures local syntax. Yarowsky [1994] has exploited this complementarity by combining the two methods using decision lists. The idea is to pool the evidence provided by the component methods, and to then solve a target problem by applying the single strongest piece of evidence, whatever type it happens to be. Yarowsky applied his method to the task of restoring missing accents in Spanish and French, and found that it outperformed both the method based on context words, and one based on local syntax. This paper takes Yarowsky's method as a starting point, and hypothesizes that further improvements can be obtained by taking into account not only the single strongest piece of evidence, but *all* the available evidence. A method is presented for doing this, based on Bayesian classifiers.

The work reported here was applied not to accent restoration, but to a related lexical disambiguation task: context-sensitive spelling correction. The task is to fix spelling errors that happen to result in valid words in the lexicon; for example:

<div align="center">

I'd like the chocolate cake for *desert.

</div>

where *dessert* was misspelled as *desert*. This goes beyond the capabilities of conventional spell checkers, which can only detect errors that result in non-words.

We start by applying a very simple method to the task, to serve as a baseline for comparison with the other methods. We then apply each of the two component methods mentioned above — context words and collocations. We try two ways of combining these components: decision lists, and Bayesian classifiers. We evaluate the above methods by comparing them with an alternative approach to spelling correction based on part-of-speech trigrams.

The sections below discuss the task of context-sensitive spelling correction, the five methods we tried for the task (baseline, two component methods, and two hybrid methods), and the evaluation. The final section draws some conclusions.

## 2 Context-sensitive spelling correction

Context-sensitive spelling correction is the problem of correcting spelling errors that result in valid words in the lexicon. Such errors can arise for a variety of reasons, including typos (e.g., *out* for *our*), homonym confusions (*there* for *their*), and usage errors (*between* for *among*). These errors are not detected by conventional spell checkers, as they only notice errors resulting in non-words.

We treat context-sensitive spelling correction as a task of word disambiguation. The ambiguity among words is modelled by *confusion sets*. A confusion set $C = \{w_1, \ldots, w_n\}$ means that each word $w_i$ in the set is ambiguous with each other word in the set. Thus if $C = \{desert, dessert\}$, then when the spelling-correction program sees an occurrence of either *desert* or *dessert* in the target document, it takes it to be ambiguous between *desert* and *dessert*, and tries to infer from the context which of the two it should be.

This treatment requires a collection of confusion sets to start with. There are several ways to obtain such a collection. One is based on finding words in the dictionary that are one typo away from each other [Mays *et al.*, 1991].[1] Another finds words that have the same or similar pronunciations. Since this was not the focus of the work reported here, we simply took (most of) our confusion sets from the list of "Words Commonly Confused" in the back of the Random House unabridged dictionary [Flexner, 1983].

A final point concerns the two types of errors a spelling-correction program can make: false negatives (complaining about a correct word), and false positives (failing to notice an error). We will make the simplifying assumption that both kinds of errors are equally bad. In practice, however, false negatives are much worse, as users get irritated by programs that badger them with bogus complaints. However, given the probabilistic nature of the methods that will be presented below, it would not be hard to modify them to take this into account. We would merely set a confidence threshold, and report a suggested correction only if the probability of the suggested word exceeds the probability of the user's original spelling by at least the threshold amount. The reason this was not done in the work reported here is that setting this confidence threshold involves a certain subjective factor (which depends on the user's "irritability threshold"). Our simplifying assumption allows us to measure performance objectively, by the single parameter of prediction accuracy.

---

[1] Constructing confusion sets in this way requires assigning each word in the lexicon its own confusion set. For instance, *cat* might have the confusion set $\{hat, car, \ldots\}$, *hat* might have $\{cat, had, \ldots\}$, and so on. We cannot use the *symmetric* confusion sets that we have adopted — where every word in the set is confusable with every other one — because the "confusable" relation is no longer transitive.

# 3 Five methods for spelling correction

This section presents a progression of five methods for context-sensitive spelling correction:

**Baseline** An indicator of "minimal competency" for comparison with the other methods
**Context words** Tests for particular words within $\pm k$ words of the ambiguous target word
**Collocations** Tests for syntactic patterns around the ambiguous target word
**Decision lists** Combines context words and collocations via decision lists
**Bayesian classifiers** Combines context words and collocations via Bayesian classifiers.

Each method will be described in terms of its operation on a single confusion set $C = \{w_1, \ldots, w_n\}$; that is, we will say how the method disambiguates occurrences of words $w_1$ through $w_n$ from the context. The methods handle multiple confusion sets by applying the same technique to each confusion set independently.

Each method involves a training phase and a test phase. The performance figures given below are based on training each method on the 1-million-word Brown corpus [Kučera and Francis, 1967] and testing it on a 3/4-million-word corpus of Wall Street Journal text [Marcus *et al.*, 1993].

## 3.1 Baseline method

The baseline method disambiguates words $w_1$ through $w_n$ by simply ignoring the context, and always guessing that the word should be whichever $w_i$ occurred most often in the training corpus. For instance, if $C = \{desert, dessert\}$, and *desert* occurred more often than *dessert* in the training corpus, then the method will predict that every occurrence of *desert* or *dessert* in the test corpus should be changed to (or left as) *desert*.

Table 1 shows the performance of the baseline method for 18 confusion sets. This collection of confusion sets will be used for evaluating the methods throughout the paper. Each line of the table gives the results for one confusion set: the words in the confusion set; the number of instances of any word in the confusion set in the training corpus and in the test corpus; the word in the confusion set that occurred most often in the training corpus; and the prediction accuracy of the baseline method for the test corpus. Prediction accuracy is the number of times the correct word was predicted, divided by the total number of test cases. For example, the members of the confusion set $\{I, me\}$ occurred 840 times in the test corpus, the breakdown being 744 *I* and 96 *me*. The baseline method predicted *I* every time, and thus was right 744 times, for a score of $744/840 = 0.886$.

Essentially the baseline method measures how accurately one can predict words using just their prior probabilities. This provides a lower bound on the performance we would expect from the other methods, which use more than just the priors.

## 3.2 Component method 1: Context words

One clue about the identity of an ambiguous target word comes from the words around it. For instance, if the target word is ambiguous between *desert* and *dessert*, and we see words like *arid*, *sand*, and *sun* nearby, this suggests that the target word should be *desert*. On the other hand, words such as *chocolate* and *delicious* in the context imply *dessert*. This observation is the basis for the method of context words. The idea is that each word $w_i$ in the confusion set will have a characteristic distribution of words that occur in its context; thus to classify an ambiguous target word, we look at the set of words around it and see which $w_i$'s distribution they most closely follow.

| Confusion set | No. of training cases | No. of test cases | Most frequent word | Baseline |
|---|---|---|---|---|
| whether, weather | 331 | 245 | whether | 0.922 |
| I, me | 6125 | 840 | I | 0.886 |
| its, it's | 1951 | 3575 | its | 0.863 |
| past, passed | 385 | 397 | past | 0.861 |
| than, then | 2949 | 1659 | than | 0.807 |
| being, begin | 727 | 449 | being | 0.780 |
| effect, affect | 228 | 162 | effect | 0.741 |
| your, you're | 1047 | 212 | your | 0.726 |
| number, amount | 588 | 429 | number | 0.627 |
| council, counsel | 82 | 83 | council | 0.614 |
| rise, raise | 139 | 301 | rise | 0.575 |
| between, among | 1003 | 730 | between | 0.538 |
| led, lead | 226 | 219 | led | 0.530 |
| except, accept | 232 | 95 | except | 0.442 |
| peace, piece | 310 | 61 | peace | 0.393 |
| there, their, they're | 5026 | 2187 | there | 0.306 |
| principle, principal | 184 | 69 | principle | 0.290 |
| sight, site, cite | 149 | 44 | sight | 0.114 |

Table 1: Performance of the baseline method for 18 confusion sets. The "Most frequent word" column gives the word in the confusion set that occurred most frequently in the training corpus. (In subsequent tables, confusion sets will be referred to by their most frequent word.) The "Baseline" column gives the prediction accuracy of the baseline system on the test corpus.

Following previous work [Gale *et al.*, 1994], we formulate the method in a Bayesian framework. The task is to pick the word $w_i$ that is most probable, given the context words $c_j$ observed within a $\pm k$-word window of the target word. The probability for each $w_i$ is calculated using Bayes' rule:

$$p(w_i|c_{-k},\ldots,c_{-1},c_1,\ldots,c_k) = \frac{p(c_{-k},\ldots,c_{-1},c_1,\ldots,c_k|w_i)\,p(w_i)}{p(c_{-k},\ldots,c_{-1},c_1,\ldots,c_k)}$$

As it stands, the likelihood term, $p(c_{-k},\ldots,c_{-1},c_1,\ldots,c_k|w_i)$, is difficult to estimate from training data — we would have to count situations in which the entire context was previously observed around word $w_i$, which raises a severe sparse-data problem. Instead, therefore, we assume that the presence of one word in the context is independent of the presence of any other word. This lets us decompose the likelihood into a product:

$$p(c_{-k},\ldots,c_{-1},c_1,\ldots,c_k|w_i) = \prod_{j\in -k,\ldots,-1,1,\ldots,k} p(c_j|w_i)$$

Gale *et al.* [1994] provide evidence that this is in fact a reasonable approximation.

We still have the problem, however, of estimating the individual $p(c_j|w_i)$ probabilities from our training corpus. The straightforward way would be to use a maximum likelihood estimate — we would count $M_i$, the total number of occurrences of $w_i$ in the training corpus, and $m_i$, the number of such occurrences for which $c_j$ occurred within $\pm k$ words, and we would then take the ratio $m_i/M_i$.[2] Unfortunately, we may not have enough training data to get an accurate estimate this way. Gale *et al.* [1994] address this problem by interpolating between two maximum-likelihood estimates: one of $p(c_j|w_i)$, and one of $p(c_j)$. The former measures the desired quantity, but is subject to inaccuracy due to sparse data; the latter provides a robust estimate, but of a potentially irrelevant quantity. Gale *et al.* interpolate between the two so as to minimize the overall inaccuracy.

We have pursued an alternative approach to the problem of estimating the likelihood terms. We start with the observation that there is no need to use *every* word in the $\pm k$-word window to discriminate among the words in the confusion set. If we do not have enough training data for a given word $c$ to accurately estimate $p(c|w_i)$ for all $w_i$, then we simply disregard $c$, and base our discrimination on other, more reliable evidence. We implement this by introducing a "minimum occurrences" threshold, $T_{min}$. It is currently set to 10. We then ignore a context word $c$ if:

$$\sum_{1 \le i \le n} m_i < T_{min} \quad \text{or} \quad \sum_{1 \le i \le n} (M_i - m_i) < T_{min}$$

where $m_i$ and $M_i$ are defined as above. In other words, $c$ is ignored if it practically never occurs within the context of any $w_i$, or if it practically always occurs within the context of every $w_i$. In the former case, we have insufficient data to measure its presence; in the latter, its absence.

Besides the reason of insufficient data, a second reason to ignore a context word is if it does not help discriminate among the words in the confusion set. For instance, if we are trying to decide between *I* and *me*, then the presence of *the* in the context probably does not help. By ignoring such words, we eliminate a source of noise in our discrimination procedure, as well as reducing storage requirements and run time. To determine whether a context word $c$ is a useful discriminator, we run a chi-square test [Fleiss, 1981] to check for an association between the presence of $c$ and the choice of word in the confusion set. If the observed association is not judged to be significant,[3] then $c$ is discarded. The significance level is currently set to 0.05.

Figure 1 pulls together the points of the preceding discussion into an outline of the method of context words. In the training phase, it identifies a list of context words that are useful for discriminating among the words in the confusion set. At run time, it estimates the probability of each word in the confusion set. It starts with the prior probabilities, and multiplies them by the likelihood of each context word from its list that appears in the $\pm k$-word window of the target word. Finally, it selects the word in the confusion set with the greatest probability.

The main parameter to tune for the method of context words is $k$, the half-width of the context window. Previous work [Yarowsky, 1994] shows that smaller values of $k$ (3 or 4) work well for resolving local syntactic ambiguities, while larger values (20 to 50) are suitable for resolving semantic ambiguities. We tried the values 3, 6, 12, and 24 on some practice confusion sets (not shown here), and found that $k = 3$ generally did best, indicating that most of the action, for our task and confusion sets, comes from local syntax. In the rest of this paper, this value of $k$ will be used.

---

[2] We are interpreting the condition "$c_j$ occurs within a $\pm k$-word window of $w_i$" as a binary feature — either it happens, or it does not. This allows us to handle context words in the same Bayesian framework as will be used later for other binary features (see Section 3.3). A more conventional interpretation is to take into account the *number* of occurrences of each $c_j$ within the $\pm k$-word window, and to estimate $p(c_j|w_i)$ accordingly. However, either interpretation is valid, as long as it is applied consistently — that is, both when estimating the likelihoods from training data, and when classifying test cases.

[3] An association is significant if the probability that it occurred by chance is low. This is not a statement about the *strength* of the association. Even a weak association may be judged significant if there are enough data to support it. Measures of the strength of association will be discussed in Section 3.4.

| Training phase |
| --- |
| (1) Propose all words as candidate context words. |
| (2) Count occurrences of each candidate context word in the training corpus. |
| (3) Prune context words that have insufficient data or are uninformative discriminators. |
| (4) Store the remaining context words (and their associated statistics) for use at run time. |
| |
| **Run time** |
| |
| (1) Initialize the probability for each word in the confusion set to its prior probability. |
| (2) Go through the list of context words that was saved during training. For each context word that appears in the context of the ambiguous target word, update the probabilities. |
| (3) Choose the word in the confusion set with the highest probability. |

Figure 1: Outline of the method of context words.

Table 2 shows the effect of varying $k$ for our usual collection of confusion sets. It can be seen that performance generally degrades as $k$ increases. The reason is that the method starts picking up spurious correlations in the training corpus. Table 4 gives some examples of the context words learned for the confusion set {*peace*, *piece*}, with $k = 24$. The context words *corps*, *united*, *nations*, etc., all imply *peace*, and appear to be plausible (although *united* and *nations* are a counterexample to our earlier assumption of independence). On the other hand, consider the context word *how*, which allegedly also implies *peace*. If we look back at the training corpus for the supporting data for this word, we find excerpts such as:

> But oh, <u>how</u> I do sometimes need just a moment of rest, and *peace* ...
> No matter <u>how</u> earnest is our quest for guaranteed *peace* ...
> <u>How</u> best to destroy your *peace* ?

There does not seem to be a necessary connection here between *how* and *peace*; the correlation is probably spurious. Although we are using a chi-square test expressly to filter out such spurious correlations, we can only expect the test to catch 95% of them (given that the significance level was set to 0.05). As mentioned above, most of the legitimate context words show up for small $k$; thus as $k$ gets large, the limited number of legitimate context words gets overwhelmed by the 5% of the spurious correlations that make it through our filter.

## 3.3 Component method 2: Collocations

The method of context words is good at capturing generalities that depend on the presence of nearby words, but not their order. When order matters, other more syntax-based methods, such as collocations and trigrams, are appropriate. In the work reported here, the method of collocations was used to capture order dependencies. A collocation expresses a pattern of syntactic elements around the target word. We allow two types of syntactic elements: words, and part-of-speech tags. Going back to the {*desert*, *dessert*} example, a collocation that would imply *desert* might be:

PREP the __

| Confusion set | Baseline | Cwords ±3 | Cwords ±6 | Cwords ±12 | Cwords ±24 |
|---|---|---|---|---|---|
| whether | 0.922 | 0.902 | 0.922 | 0.927 | 0.922 |
| I | 0.886 | 0.914 | 0.893 | 0.883 | 0.851 |
| its | 0.863 | 0.862 | 0.795 | 0.743 | 0.702 |
| past | 0.861 | 0.861 | 0.849 | 0.801 | 0.743 |
| than | 0.807 | 0.931 | 0.901 | 0.896 | 0.855 |
| being | 0.780 | 0.791 | 0.795 | 0.793 | 0.755 |
| effect | 0.741 | 0.747 | 0.741 | 0.759 | 0.716 |
| your | 0.726 | 0.816 | 0.783 | 0.774 | 0.736 |
| number | 0.627 | 0.646 | 0.622 | 0.636 | 0.639 |
| council | 0.614 | 0.639 | 0.614 | 0.602 | 0.614 |
| rise | 0.575 | 0.575 | 0.575 | 0.585 | 0.498 |
| between | 0.538 | 0.759 | 0.697 | 0.671 | 0.586 |
| led | 0.530 | 0.530 | 0.530 | 0.521 | 0.557 |
| except | 0.442 | 0.695 | 0.526 | 0.516 | 0.558 |
| peace | 0.393 | 0.754 | 0.705 | 0.574 | 0.574 |
| there | 0.306 | 0.726 | 0.623 | 0.557 | 0.466 |
| principle | 0.290 | 0.290 | 0.290 | 0.290 | 0.435 |
| sight | 0.114 | 0.455 | 0.250 | 0.364 | 0.318 |
| Avg no. of context words | | 27.9 | 36.9 | 55.9 | 92.9 |

Table 2: Performance of the method of context words as a function of $k$, the half-width of the context window. The bottom line of the table shows the number of context words learned, averaged over all confusion sets, also as a function of $k$.

This collocation would match the sentences:

> Travelers entering from the *desert* were confounded...
> ...along with some guerrilla fighting in the *desert*.
> ...two ladies who lay pinkly nude beside him in the *desert* ...

Matching part-of-speech tags (here, PREP) against the sentence is done by first tagging each word in the sentence with its *set* of possible part-of-speech tags, obtained from a dictionary. For instance, *walk* has the tag set {NS, V}, corresponding to its use as a singular noun and as a verb.[4] For a tag to match a word, the tag must be a member of the word's tag set. The reason we use tag *sets*, instead of running a tagger on the sentence to produce unique tags, is that taggers need to look at all words in the sentence, which is impossible when the target word is taken to be ambiguous (but see the trigram method in Section 4).

The method of collocations was implemented in much the same way as the method of context words. The idea is to discriminate among the words $w_i$ in the confusion set by identifying the collocations that tend to occur around each $w_i$. An ambiguous target word is then classified by finding all collocations that match its context. Each collocation provides some degree of evidence for each word in the confusion set. This evidence is combined using Bayes' rule. In the end, the $w_i$ with the highest probability, given the evidence, is selected.

---

[4] Our tag inventory contains 40 tags, and includes the usual categories for determiners, nouns, verbs, modals, etc., a few specialized tags (for *be*, *have*, and *do*), and a dozen compound tags (such as V+PRO for *let's*).

A new complication arises for collocations, however, in that collocations, unlike context words, cannot be assumed independent. Consider, for example, the following collocations for *desert*:

PREP the __
in the __
the __

These collocations are highly interdependent — we will say they *conflict*. To deal with this problem, we invoke our earlier observation that there is no need to use all the evidence. If two pieces of evidence conflict, we simply eliminate one of them, and base our decision on the rest of the evidence. We identify conflicts by the heuristic that two collocations conflict iff they overlap. The overlapping portion is the factor they have in common, and thus represents their lack of independence. This is only a heuristic because we could imagine collocations that do *not* overlap, but still conflict. Note, incidentally, that there can be at most two non-conflicting collocations for any decision — one matching on the left-hand side of the target word, and one on the right.

Having said that we resolve conflicts between two collocations by eliminating one of them, we still need to specify which one. Our approach is to assign each one a *strength*, just as Yarowsky [1994] does in his hybrid method, and to eliminate the one with the lower strength. This preserves the strongest non-conflicting evidence as the basis for our answer. The strength of a collocation reflects its reliability for decision-making; a further discussion of strength is deferred to Section 3.4.

Figure 2 ties together the preceding discussion into an outline of the method of collocations. The method is described in terms of "features" rather than "collocations" to reflect its full generality; the features could be context words as well as collocations. In fact, the method subsumes the method of context words — it does everything that method does, and resolves conflicts among its features as well. To facilitate the conflict resolution, it sorts the features by decreasing strength.

Like the method of context words, the method of collocations has one main parameter to tune: $\ell$, the maximum number of syntactic elements in a collocation. Since the number of collocations grows exponentially with $\ell$, it was only practical to vary $\ell$ from 1 to 3. We tried this on some practice confusion sets, and found that all values of $\ell$ gave roughly comparable performance. We selected $\ell = 2$ to use from here on, as a compromise between reducing the expressive power of collocations (with $\ell = 1$) and incurring a high computational cost (with $\ell = 3$).

Table 3 shows the results of varying $\ell$ for the usual confusion sets. There is no clear winner; each value of $\ell$ did best for certain confusion sets. Table 5 gives examples of the collocations learned for $\{peace, piece\}$ with $\ell = 2$. A good deal of redundancy can be seen among the collocations. There is also some redundancy between the collocations and the context words of the previous section (e.g., for *corps*). Many of the collocations at the end of the list appear to be overgeneral and irrelevant.

## 3.4   Hybrid method 1: Decision lists

Yarowsky [1994] pointed out the complementarity between context words and collocations: context words pick up those generalities that are best expressed in an order-independent way, while collocations capture order-dependent generalities. Yarowsky proposed decision lists as a way to get the best of both methods. The idea is to make one big list of all features — in this case, context words and collocations. The features are sorted in order of decreasing strength, where the strength of a feature reflects its reliability for decision-making. An ambiguous target word is then classified by running down the list and matching each feature against the target context. The first feature that

| Training phase |
| --- |
| (1)  Propose all possible features as candidate features. |
| (2)  Count occurrences of each candidate feature in the training corpus. |
| (3)  Prune features that have insufficient data or are uninformative discriminators. |
| **(3.5)**  **Sort the remaining features in order of decreasing strength.** |
| (4)  Store the list of features (and their associated statistics) for use at run time. |

<table>
<tr><td colspan="2"><u>Run time</u></td></tr>
<tr><td>(1)</td><td>Initialize the probability for each word in the confusion set to its prior probability.</td></tr>
<tr><td>(2)</td><td>Go through the sorted list of features that was saved during training. For each feature that matches the context of the ambiguous target word, <b>and does not conflict with a feature accepted previously,</b> update the probabilities.</td></tr>
<tr><td>(3)</td><td>Choose the word in the confusion set with the highest probability.</td></tr>
</table>

Figure 2: Outline of the method of collocations. Differences from the method of context words are highlighted in boldface. The method is described in terms of "features" rather than "collocations" to reflect its full generality.

matches is used to classify the target word. Yarowsky [1994] describes further refinements, such as detecting and pruning features that make a zero or negative contribution to overall performance.

The method of decision lists, as just described, is almost the same as the method for collocations in Figure 2, where we take "features" in that figure to include both context words and collocations. The main difference is that during evidence gathering (step (2) at run time), decision lists terminate after matching the first feature. This obviates the need for resolving conflicts between features.

Given that decision lists base their answer for a problem on the single strongest feature, their performance rests heavily on how the strength of a feature is defined. Yarowsky [1994] used the following metric to calculate the strength of a feature $f$:

$$reliability(f) = \text{abs}\left(\log\left(\frac{p(w_1|f)}{p(w_2|f)}\right)\right)$$

This is for the case of a confusion set of two words, $w_1$ and $w_2$. It can be shown that this metric produces the identical ranking of features as the following somewhat simpler metric, provided $p(w_i|f) > 0$ for all $i$:[5]

$$reliability'(f) = \max_i p(w_i|f)$$

As an example of using the metric, suppose $f$ is the context word *arid*, and suppose that *arid* co-occurs 10 times with *desert* and 1 time with *dessert* in the training corpus. Then $reliability'(f) = \max(10/11, 1/11) = 10/11 = 0.909$. This value measures the extent to which the presence of the feature is unambiguously correlated with one particular $w_i$. It can be thought of as the feature's *reliability* at picking out that $w_i$ from the others in the confusion set.

---

[5]In fact, we guarantee that this inequality holds by performing smoothing before calculating strength. We smooth the data by adding 1 to the count of how many times each feature was observed for each $w_i$.

| Confusion set | Baseline | Collocs $\leq 1$ | Collocs $\leq 2$ | Collocs $\leq 3$ |
|---|---|---|---|---|
| whether | 0.922 | 0.939 | 0.931 | 0.931 |
| I | 0.886 | 0.979 | 0.981 | 0.980 |
| its | 0.863 | 0.943 | 0.945 | 0.950 |
| past | 0.861 | 0.919 | 0.909 | 0.909 |
| than | 0.807 | 0.966 | 0.965 | 0.966 |
| being | 0.780 | 0.853 | 0.853 | 0.842 |
| effect | 0.741 | 0.821 | 0.821 | 0.821 |
| your | 0.726 | 0.877 | 0.887 | 0.887 |
| number | 0.627 | 0.646 | 0.646 | 0.681 |
| council | 0.614 | 0.663 | 0.639 | 0.639 |
| rise | 0.575 | 0.807 | 0.807 | 0.807 |
| between | 0.538 | 0.699 | 0.730 | 0.733 |
| led | 0.530 | 0.849 | 0.840 | 0.863 |
| except | 0.442 | 0.800 | 0.789 | 0.789 |
| peace | 0.393 | 0.869 | 0.869 | 0.852 |
| there | 0.306 | 0.911 | 0.932 | 0.932 |
| principle | 0.290 | 0.841 | 0.812 | 0.812 |
| sight | 0.114 | 0.341 | 0.318 | 0.318 |
| Avg no. of collocations | | 33.9 | 263.1 | 985.4 |

Table 3: Performance of the method of collocations as a function of $\ell$, the maximum length of a collocation. The bottom line of the table shows the number of collocations learned, averaged over all confusion sets, also as a function of $\ell$.

One peculiar property of the reliability metric is that it ignores the prior probabilities of the words in the confusion set. For instance, in the *arid* example, it would award the same high score even if the total number of occurrences of *desert* and *dessert* in the training corpus were 50 and 5, respectively — in which case *arid*'s performance of 10/11 would be exactly what one would expect by chance, and therefore hardly impressive. Besides the reliability metric, therefore, we also considered an alternative metric: the uncertainty coefficient of $x$, denoted $U(x|y)$ [Press *et al.*, 1988, p.501]. $U(x|y)$ measures how much additional information we get about the presence of the feature by knowing the choice of word in the confusion set.[6] $U(x|y)$ is calculated as follows:

$$U(x|y) = \frac{H(x) - H(x|y)}{H(x)}$$
$$H(x) = -p(f) \ln p(f) - p(\neg f) \ln p(\neg f)$$
$$H(x|y) = -\sum_i p(w_i) \left( p(f|w_i) \ln p(f|w_i) + p(\neg f|w_i) \ln p(\neg f|w_i) \right)$$

The probabilities are calculated for the population consisting of all occurrences in the training corpus of any $w_i$. For instance, $p(f)$ is the probability of feature $f$ being present within this

---

[6]This definition may seem backwards, but is appropriate for use on the right-hand side of Bayes' rule, where the choice of word in the confusion set is the "given".

| Context word | *peace* | *piece* |
|---|---|---|
| corps | 49 | 1 |
| peace | 41 | 1 |
| united | 20 | 0 |
| nations | 15 | 0 |
| our | 27 | 1 |
| heart | 12 | 0 |
| justice | 12 | 0 |
| state | 12 | 0 |
| american | 11 | 0 |
| aid | 11 | 0 |
| international | 11 | 0 |
| women | 10 | 0 |
| war | 20 | 1 |
| world | 40 | 3 |
| piece | 1 | 15 |
| over | 1 | 14 |
| must | 11 | 1 |
| great | 11 | 1 |
| under | 10 | 1 |
| how | 10 | 1 |
| ⋮ | | |
| two | 5 | 12 |
| for | 83 | 38 |
| about | 4 | 9 |
| every | 4 | 9 |
| little | 5 | 10 |
| long | 6 | 11 |
| one | 14 | 23 |
| the | 179 | 113 |
| so | 9 | 14 |
| ; | 16 | 22 |
| Total occurrences | 184 | 126 |

Table 4: Excerpts from the list of 43 context words learned for {*peace*, *piece*} with $k = 24$. Each line gives a context word, and the number of *peace* and *piece* occurrences for which that context word occurred within $\pm k$ words. The last line of the table gives the total number of occurrences of *peace* and *piece* in the training corpus.

| Collocation | *peace* | *piece* |
|---|---|---|
| ___ corps | 47 | 0 |
| DET ___ corps | 32 | 0 |
| ADV ___ corps | 28 | 0 |
| the ___ corps | 27 | 0 |
| ___ and | 22 | 0 |
| ___ of NS | 2 | 60 |
| the ___ NS | 37 | 1 |
| a ___ PREP | 1 | 35 |
| PREP ___ of | 1 | 34 |
| a ___ of | 1 | 34 |
| for ___ | 16 | 0 |
| ___ and NS | 16 | 0 |
| DET ___ NP | 32 | 1 |
| NS ___ of | 2 | 45 |
| ___ corps NS | 14 | 0 |
| PREP ___ CONJ | 14 | 0 |
| the ___ NP | 27 | 1 |
| V CONJ ___ | 13 | 0 |
| ___ NS PUNC | 13 | 0 |
| ___ of V | 1 | 25 |
| ⋮ | | |
| CONJ ADJ ___ | 4 | 9 |
| the NS ___ | 4 | 9 |
| NS ADJ ___ | 13 | 26 |
| ADV NS ___ | 12 | 23 |
| PREP NS ___ | 17 | 31 |
| ADV ___ PREP | 12 | 22 |
| ADJ ADJ ___ | 9 | 14 |
| NS ___ | 62 | 79 |
| ADJ ___ | 46 | 54 |
| NS NS ___ | 29 | 32 |
| Total occurrences | 184 | 126 |

Table 5: Excerpts from the sorted list of 98 collocations learned for {*peace*, *piece*} with $\ell = 2$. Each line gives a collocation, and the number of *peace* and *piece* occurrences it matched. The last line of the table gives the total number of occurrences of *peace* and *piece* in the training corpus.

49

population. Applying the $U(x|y)$ metric to the *arid* example, the value returned now depends on the number of occurrences of *desert* and *dessert* in the training corpus. If these numbers are 50 and 5, then $U(x|y) = 0.0$, reflecting the uninformativeness of the *arid* feature in this situation. If instead the numbers are 50 and 500, then $U(x|y) = 0.402$, indicating *arid*'s better-than-chance ability to pick out *desert* (10 out of 50 occurrences) over *dessert* (1 out of 500 occurrences).

To compare the two strength metrics, we tried both on some practice confusion sets. Sometimes one metric did substantially better, sometimes the other. In the balance, the reliability metric seemed to give higher performance. This metric is therefore the one that will be used from here on. It was also used for all experiments involving the method of collocations.

Table 6 shows the performance of decision lists with each metric for the usual confusion sets. As with the practice confusion sets, we see sometimes dramatic performance differences between the two metrics, and no clear winner. For instance, for $\{I, me\}$, the reliability metric did better than $U(x|y)$ (0.980 versus 0.808); whereas for $\{between, among\}$, it did worse (0.659 versus 0.800). Further research is needed to understand the circumstances under which each metric performs best.

Focusing for now on the reliability metric, Table 6 shows that the method of decision lists does, by and large, accomplish what it set out to do — namely, outperform either component method alone. There are, however, a few cases where it falls short; for instance, for $\{between, among\}$, decision lists score only 0.659, compared with 0.759 for context words and 0.730 for collocations.[7] We believe that the problem lies in the strength metric: because decision lists make their judgements based on a single piece of evidence, their performance is very sensitive to the metric used to select that piece of evidence. But as the reliability and $U(x|y)$ metrics indicate, it is not completely clear how the metric should be defined. This problem is addressed in the next section.

## 3.5 Hybrid method 2: Bayesian classifiers

The previous section confirmed that decision lists are effective at combining two complementary methods — context words and collocations. In doing the combination, however, decision lists look only at the single strongest piece of evidence for a given problem. We hypothesize that even better performance can be obtained by taking into account *all* available evidence. This section presents a method of doing this based on Bayesian classifiers.

Like decision lists, the Bayesian method starts with a list of all features, sorted by decreasing strength. It classifies an ambiguous target word by matching each feature in the list in turn against the target context. Instead of stopping at the first matching feature, however, it traverses the entire list, combining evidence from all matching features, and resolving conflicts where necessary.

This method is essentially the same as the one for collocations (see Figure 2), except that it uses context words as well as collocations for the features. The only new wrinkle is in checking for conflicts between features (in step (2) at run time), as there are now two kinds of features to consider. If both features are context words, we say the features never conflict (as in the method of context words). If both features are collocations, we say they conflict iff they overlap (as in the method of collocations). The new case is if one feature is a context word, and the other is a collocation. Consider, for example, the context word *walk*, and the following collocations:

(1)       __ walk
(2)       V
(3)   CONJ __ PREP

---

[7]If we use the $U(x|y)$ metric instead, then decision lists fall down on different examples; e.g., $\{its, it's\}$.

| Confusion set | Baseline | Cwords ±3 | Collocs ≤ 2 | Dlist Rely | Dlist $U(x\|y)$ |
|---|---|---|---|---|---|
| whether | 0.922 | 0.902 | 0.931 | 0.935 | 0.829 |
| I | 0.886 | 0.914 | 0.981 | 0.980 | 0.808 |
| its | 0.863 | 0.862 | 0.945 | 0.931 | 0.805 |
| past | 0.861 | 0.861 | 0.909 | 0.932 | 0.892 |
| than | 0.807 | 0.931 | 0.965 | 0.967 | 0.961 |
| being | 0.780 | 0.791 | 0.853 | 0.842 | 0.933 |
| effect | 0.741 | 0.747 | 0.821 | 0.821 | 0.654 |
| your | 0.726 | 0.816 | 0.887 | 0.868 | 0.896 |
| number | 0.627 | 0.646 | 0.646 | 0.629 | 0.667 |
| council | 0.614 | 0.639 | 0.639 | 0.627 | 0.651 |
| rise | 0.575 | 0.575 | 0.807 | 0.804 | 0.827 |
| between | 0.538 | 0.759 | 0.730 | 0.659 | 0.800 |
| led | 0.530 | 0.530 | 0.840 | 0.840 | 0.840 |
| except | 0.442 | 0.695 | 0.789 | 0.789 | 0.726 |
| peace | 0.393 | 0.754 | 0.869 | 0.852 | 0.836 |
| there | 0.306 | 0.726 | 0.932 | 0.914 | 0.906 |
| principle | 0.290 | 0.290 | 0.812 | 0.812 | 0.841 |
| sight | 0.114 | 0.455 | 0.318 | 0.432 | 0.568 |

Table 6: Performance of decision lists with the reliability and $U(x|y)$ strength metrics.

To some extent, all of these collocations conflict with *walk*. Collocation (1) is the most blatant case; if it matches the target context, this logically *implies* that the context word *walk* will match. If collocation (2) matches, this guarantees that one of the possible tags of *walk* will be present nearby the target word, thereby elevating the probability that *walk* will match within ±*k* words. If collocation (3) matches, this guarantees that there are two positions nearby the target word that are incompatible with *walk*, thereby *reducing* the probability that *walk* will match. If we were to treat all of these cases as conflicts, we would end up losing a great deal of (potentially useful) evidence. Instead, we adopt the more relaxed policy of only flagging the most egregious conflicts — here, the one between collocation (1) and *walk*. In general, we will say that a collocation and a context word conflict iff the collocation contains an explicit test for the context word.

Table 7 compares all methods covered so far — baseline, two component methods, and two hybrid methods. (A sixth method, trigrams, is included as well — it will be discussed in Section 4.) The table shows that the Bayesian hybrid method does at least as well as the previous four methods for almost every confusion set. Occasionally it scores slightly less than collocations; this appears to be due to some averaging effect where noisy context words are dragging it down. Occasionally too it scores less than decision lists, but never by much; on the whole, it yields a modest but consistent improvement, and in the case of {*between, among*}, a sizable improvement. We believe the improvement is due to considering all of the evidence, rather than just the single strongest piece, which makes the method more robust to inaccurate judgements about which piece of evidence is "strongest".

| Confusion set | Baseline | Cwords ±3 | Collocs ≤ 2 | Dlist Rely | Bayes Rely | Trigrams |
|---|---|---|---|---|---|---|
| whether | 0.922 | 0.902 | 0.931 | 0.935 | 0.935 | 0.873 |
| I | 0.886 | 0.914 | 0.981 | 0.980 | 0.985 | 0.985 |
| its | 0.863 | 0.862 | 0.945 | 0.931 | 0.942 | 0.965 |
| past | 0.861 | 0.861 | 0.909 | 0.932 | 0.924 | 0.955 |
| than | 0.807 | 0.931 | 0.965 | 0.967 | 0.973 | 0.780 |
| being | 0.780 | 0.791 | 0.853 | 0.842 | 0.869 | 0.978 |
| effect | 0.741 | 0.747 | 0.821 | 0.821 | 0.827 | 0.975 |
| your | 0.726 | 0.816 | 0.887 | 0.868 | 0.901 | 0.958 |
| number | 0.627 | 0.646 | 0.646 | 0.629 | 0.662 | 0.636 |
| council | 0.614 | 0.639 | 0.639 | 0.627 | 0.639 | 0.651 |
| rise | 0.575 | 0.575 | 0.807 | 0.804 | 0.807 | 0.574 |
| between | 0.538 | 0.759 | 0.730 | 0.659 | 0.786 | 0.538 |
| led | 0.530 | 0.530 | 0.840 | 0.840 | 0.840 | 0.909 |
| except | 0.442 | 0.695 | 0.789 | 0.789 | 0.811 | 0.695 |
| peace | 0.393 | 0.754 | 0.869 | 0.852 | 0.852 | 0.393 |
| there | 0.306 | 0.726 | 0.932 | 0.914 | 0.916 | 0.961 |
| principle | 0.290 | 0.290 | 0.812 | 0.812 | 0.812 | 0.609 |
| sight | 0.114 | 0.455 | 0.318 | 0.432 | 0.455 | 0.250 |

Table 7: Performance of six methods for context-sensitive spelling correction.

# 4  Evaluation

While the previous section demonstrated that the Bayesian hybrid method does better than its components, we would still like to know how it compares with alternative methods. We looked at a method based on part-of-speech trigrams, developed and implemented by Schabes [1995].

Schabes's method can be viewed as performing an abductive inference: given a sentence containing an ambiguous word, it asks which choice $w_i$ for that word would best explain the observed sequence of words in the sentence. It answers this question by substituting each $w_i$ in turn into the sentence. The $w_i$ that produces the highest-probability sentence is selected. Sentence probabilities are calculated using a part-of-speech trigram model.

We tried Schabes's method on the usual confusion sets; the results are in the last column of Table 7. It can be seen that trigrams and the Bayesian hybrid method each have their better moments. Trigrams are at their worst when the words in the confusion set have the same part of speech. In this case, trigrams can distinguish between the words only by their prior probabilities — this follows from the way the method calculates sentence probabilities. Thus, for {between, among}, for example, where both words are prepositions, trigrams score the same as the baseline method. In such cases, the Bayesian hybrid method is clearly better. On the other hand, when the words in the confusion set have different parts of speech — as in, for example, {there, their, they're} — trigrams are often better than the Bayesian method. We believe this is because trigrams look not just at a few words on either side of the target word, but at the part-of-speech sequence of the whole sentence. This analysis indicates a complementarity between trigrams and Bayes, and suggests a combination in which trigrams would be applied first, but if trigrams determine that the words in the confusion set have the same part of speech for the sentence at issue, then the sentence would be passed to the Bayesian method. This is a research direction we plan to pursue.

## 5    Conclusion

The work reported here builds on Yarowsky's use of decision lists to combine two component methods — context words and collocations. Decision lists pool the evidence from the two methods, and solve a target problem by applying the single strongest piece of evidence, whichever type that happens to be. This paper investigated the hypothesis that even better performance can be obtained by basing decisions on not just the single strongest piece of evidence, but on all available evidence. A method for doing this, based on Bayesian classifiers, was presented. It was applied to the task of context-sensitive spelling correction, and was found to outperform the component methods as well as decision lists. A comparison of the Bayesian hybrid method with Schabes's trigram-based method suggested a further combination in which trigrams would be used when the words in the confusion set had different parts of speech, and the Bayesian method would be used otherwise. This is a direction we plan to pursue in future research.

## Acknowledgements

We would like to thank Bill Freeman, Yves Schabes, Emmanuel Roche, and Jacki Golding for helpful and enjoyable discussions on the work reported here.

## References

Joseph L. Fleiss. *Statistical Methods for Rates and Proportions.* John Wiley and Sons, 1981.

Stuart Berg Flexner, editor. *Random House Unabridged Dictionary.* Random House, New York, 1983. Second edition.

William A. Gale, Kenneth W. Church, and David Yarowsky. Discrimination decisions for 100,000-dimensional spaces. In *Current Issues in Computational Linguistics: In Honour of Don Walker,* pages 429–450. Kluwer Academic Publishers, 1994.

H. Kučera and W. N. Francis. *Computational Analysis of Present-Day American English.* Brown University Press, Providence, RI, 1967.

Mitchell P. Marcus, Beatrice Santorini, and Mary Ann Marcinkiewicz. Building a large annotated corpus of English: The Penn Treebank. *Computational Linguistics,* 19(2):313–330, June 1993.

Eric Mays, Fred J. Damerau, and Robert L. Mercer. Context based spelling correction. *Information Processing & Management,* 27(5):517–522, 1991.

William Press, Brian Flannery, Saul Teukolsky, and William Vetterling. *Numerical Recipes in C: The Art of Scientific Computing.* Cambridge University Press, New York, 1988. Reprinted twice.

Yves Schabes. Technical report, Mitsubishi Electric Research Laboratories, 1995. Forthcoming.

David Yarowsky. A comparison of corpus-based techniques for restoring accents in Spanish and French text. In *Proceedings of the 2nd Annual Workshop on Very Large Corpora,* Kyoto, 1994.